# Graph Attention Neural Network for Botnet Detection: Evaluating Autoencoder, VAE and PCA-Based Dimension Reduction

Hassan Wasswa[1], Hussein Abbass[1], Timothy Lynar[1]
[1]School of Systems and Computing, University of New South Wales, Canberra, ACT, Australia
h.wasswa@unsw.edu.au, h.abbass@unsw.edu.au, t.lynar@unsw.edu.au

*Abstract*—With the rise of IoT-based botnet attacks, researchers have explored various learning models for detection, including traditional machine learning, deep learning, and hybrid approaches. A key advancement involves deploying attention mechanisms to capture long-term dependencies among features, significantly improving detection accuracy. However, most models treat attack instances independently, overlooking inter-instance relationships. Graph Neural Networks (GNNs) address this limitation by learning an embedding space via iterative message passing where similar instances are placed closer based on node features and relationships, enhancing classification performance. To further improve detection, attention mechanisms have been embedded within GNNs, leveraging both long-range dependencies and inter-instance connections. However, transforming the high dimensional IoT attack datasets into a graph structured dataset poses challenges, such as large graph structures leading computational overhead. To mitigate this, this paper proposes a framework that first reduces dimensionality of the NetFlow-based IoT attack dataset before transforming it into a graph dataset. We evaluate three dimension reduction techniques—Variational Autoencoder (VAE-encoder), classical autoencoder (AE-encoder), and Principal Component Analysis (PCA)—and compare their effects on a Graph Attention neural network (GAT) model for botnet attack detection.

*Index Terms*—Autoencoder, Dimension reduction, Graph neural networks, Graph attention network, Variational autoencoder

## I. Introduction

Due to the increasing prevalence of IoT-based botnet attacks, improving the efficiency of learning models for detecting such threats has become a central focus for IoT security researchers and practitioners. Several studies have explored traditional machine learning techniques [1]– [4], advanced deep learning architectures [5]–[7], and hybrid approaches to enhance network security, particularly IoT botnet detection [1], [2], [8]. A state-of-the-art strategy that has recently gained significant attention involves embedding attention mechanisms to capture long-term dependencies among the features of input instances. Studies such as [9]–[11] have implemented attention-based models for IoT botnet detection, achieving near-perfect performance.

However, the learning process of these models typically treats each attack instance as an independent entity, disregarding the inter-relationships between different attack instances. To address this limitation, researchers are increasingly focusing on the development of Graph Neural Networks (GNNs) models for attack detection [12]–[16]. The training of these networks is designed to produce embeddings in which similar instances are positioned closer together based on learned node features and their relationships with neighboring nodes—normal instances are clustered together, while attack instances are grouped separately—thus enabling more effective flow-based traffic classification. To further improve the detection performance of GNN-based models, state-of-the-art techniques have integrated attention mechanisms within GNN architectures. This integration allows the detection model to leverage both the long-range feature dependencies captured by attention mechanisms and the inter-instance relationships learned by GNNs, leading to more robust and accurate IoT botnet detection.

Transforming flow-based IoT botnet attack datasets often involves representing each instance as a node, with edges established using metrics like Euclidean distance or cosine similarity. However, the additional edge information, combined with the high dimensional nature of each instance, can result in a large graph, making the learning process computationally expensive in terms of memory and processing power while also slowing down inference. To tackle this challenge, we propose a framework that begins by reducing the dimension of NetFlow-based IoT traffic before transforming it into a graph-structured dataset. We evaluate three dimension reduction techniques—using the encoder of a classic autoencoder (AE-encoder), using the encoder of a Variational autoencoder (VAE-encoder), and using Principal Component Analysis (PCA)—and evaluate the impact of each method on the performance of a Graph Attention Neural Network (GAT) model for botnet attack detection.

The rest of this paper is organized as follows. Section II presents prior work integrating the graph neural network model with the attention mechanism for anomaly detection. This is followed by section III where a detailed description of the framework used in this work is presented, together with a brief description of the dataset and the various techniques used. In section IV we present and discuss the findings of this study followed by the conclusion in section V.

## II. Related Work

Several recent studies have introduced hybrid approaches that integrate attention-based transformer architectures with graph neural networks (GNNs) for anomaly detection. These methods leverage the strengths of both architectures—GNNs for capturing intricate node-edge relationships and transformers for modeling long-range feature dependencies—leading to improved detection performance.

For instance, [17] proposed a model that combines a GNN with a transformer-based architecture for intrusion detection. In their approach, the GNN was optimized to capture complex structural relationships between nodes and edges, while the transformer component enhances the model's ability to detect anomalies by leveraging its capability to process long-range dependencies.

Similarly, the work in study [18] developed a framework that integrates a graph convolutional neural network with an attention mechanism to identify relationships and coordinated behaviors within IoT devices' graph-structured data. This approach enhances botnet detection by effectively modeling device interactions and uncovering hidden patterns indicative of malicious activity.

Another notable study, [19], focused on attack detection in electric vehicle (EV) charging stations. Their method involved constructing graph representations from hardware logs and utilizing a GNN to model correlations among input features. Additionally, they employ an attention-based transformer to capture complex feature interactions, ultimately improving attack detection accuracy.

Furthermore, [20] introduced AJSAGE, a framework that builds upon the GNN-based GraphSAGE model and incorporates an attention mechanism. This enhancement improves the detection of abnormal traffic nodes within graph-structured network attack data, making it more effective in identifying sophisticated cyber threats.

These studies demonstrate the growing interest in hybrid GNN-transformer models for anomaly detection, showcasing their potential in diverse applications such as intrusion detection, botnet identification, and cybersecurity in IoT environments.

### III. PROPOSED APPROACH

The proposed framework begins by projecting the high-dimensional attack dataset into a low-dimensional representation. Three dimension reduction techniques are evaluated in this phase—utilizing the AE-encoder, VAE-encoder, and PCA. The resulting low-dimensional dataset is subsequently transformed into a graph-structured representation. To achieve this, each traffic instance is treated as a node, while edges are established using the *k*-Nearest Neighbors (KNN) algorithm. Two metrics—Euclidean distance and cosine similarity—are employed to determine the neighborhood relationships be- tween instances during graph data construction. The resulting graph data is then utilized to train a Graph Attention Neural Network (GAT) model for traffic classification. A visual representation of the proposed framework is provided in Fig. 1.

#### A. Autoencoder

An autoencoder (AE) is a deep neural network architecture made up of two core components—an encoder model that projects the high-dimensional training features to a low-dimensional latent representation and a decoder model that reconstructs the high-dimensional training features from the latent space representation. For dimension reduction, the high dimensional training dataset is transformed into its low dimensional latent space representation, and the GAT model trained on it. Several studies including [21] have utilized AEs for dimension reduction. However, standard AEs lack explicit regularization in their latent space, as the training objective focuses solely on minimizing the reconstruction error. Consequently, the latent space may assume arbitrary shapes, leading to sparse representations that hinder classification performance and generalization. Additionally, the absence of regularization reduces interpretability, making it difficult to derive meaningful insights from the encoded features.

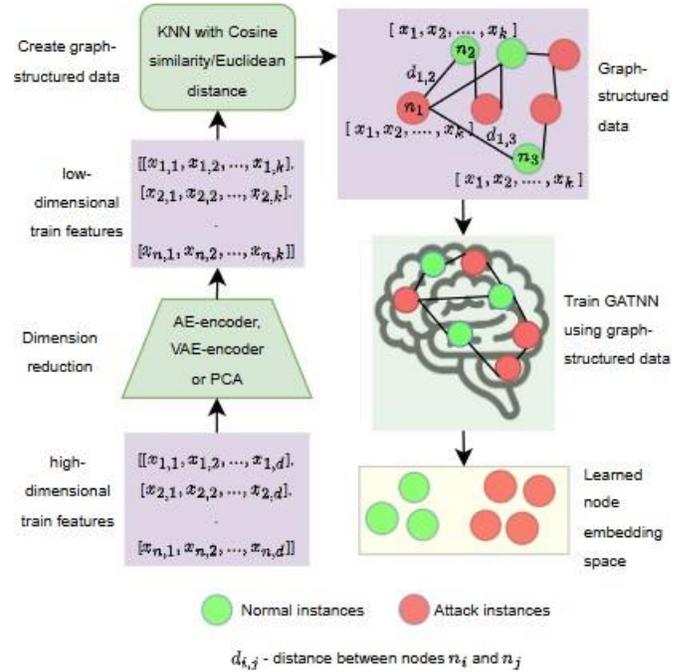

Fig. 1: Detection framework

#### B. Variational Autoencoder

The introduction of Variational Auto-encoders (VAE) [22] aimed to address the lack of regularization in the latent space, a common issue with a classical AE. A VAE by employing Bayesian variational inference to learn parameters for both the encoder and decoder. It approximates the true distribution of a dataset, $X$, by introducing a latent variable, $z$. This approach enables the learning of the joint distribution, $p_\psi(x, z)$, between $x$ and $z$ through estimating a set of parameters, $\psi$. Consequently, the marginal distribution, $p_\psi(x)$, can be estimated.

Since calculating $p_\psi(x)$ is intractable, VAE approximates $p_\psi(x)$ by assuming a simpler distribution for the latent variable $z$, treating $p_\psi(x|z)$ as $p_\psi(z|x)$. This approximation facilitates the learning of parameters $\beta$ for a surrogate network $q_\beta(z|x)$, that estimates $p_\psi(z|x)$ and allows for the calculation of the expected value of $log p_\psi(x)$ over $\beta$ as shown in Eq. 1.

$$E_{z \sim q_\beta(z|x)} \log p_\psi(x) = L(\psi, \beta, z) + D_{KL}(q_\beta(z|x)||p_\psi(z|x)) \quad (1)$$

Where $L(\psi, \beta, z)$ represents the Evidence Lower Bound (ELBO), and $D_{KL}(q_\beta(z|x)||p_\psi(z|x))$ stands for the Kullback Leibler divergence, ($D_{KL}$), between the approximate and true posterior distributions. a VAE aims to learn the parameters, $\beta$, which correspond to the decoder model's weights that approximate the posterior distribution, $q_\beta(z|x)$. Its training is focused on maximizing the ELBO and, consequently, minimizing $D_{KL}$, since the $E_{z \sim q_\beta(z|x)} \log p_\psi(x)$ is constant for a given distribution.

### C. Principal Component Analysis (PCA)

Principal Component Analysis (PCA) is a linear technique for dimension reduction that operates by computing eigenvalues and identifying eigenvectors that indicate directions of maximum variance, known as principal components. In the domain of attack detection, PCA has been applied in various studies, including [23], [24]. However, PCA is blind to the target variable categories, often yields suboptimal outcomes, requires manual determination of the explained variance threshold, and faces challenges when applied to datasets with intricate nonlinear structures [25].

### D. Dataset

The CICIoT2022 dataset [26] was used in this study. This dataset was collected over 30 days and comprises traffic instances from 60 IoT devices spanning categories such as home automation, cameras, and audio systems, utilizing communication protocols like Zigbee, Z-Wave, and WiFi. Feature extraction from the released .pcap files was performed using the revised version of the CICFlowMeter tool, yielding 84 independent features and over 3.2 million NetFlow instances. The dataset includes five class labels, derived from directory names, with the following distribution: {"Normal": 2,616,853 (80.870%), "HTTP flood": 554,316 (17.130%), "TCP flood": 45,884 (1.418%), "Brute force": 12,257 (0.379%), "UDP flood": 6,561 (0.203%)}.

### E. Graph Neural Networks

Deep learning models like CNNs and RNNs excel with structured data such as sequences, sentences, and images, but struggle with complex graph-structured data like atomic structures or network topologies. Graph Neural Networks (GNNs) address this limitation by processing graph-structured data, where nodes represent entities (e.g., atoms) and edges capture relationships (e.g., bonds). Initially proposed in [27], GNNs extend neural networks to effectively model and analyze interconnected data. This approach is especially useful for tasks where understanding the relationships between entities, such as in molecular graphs, is crucial. GNNs enable deeper insights into data with intricate and dynamic structures, making them ideal for applications in materials science, molecular biology, and NetFlow analysis.

In the context NetFlow analysis, each node in a graph represents a traffic instance, with edges defining relationships like cosine similarity or IP address connections. A GNN analyzes these nodes and their neighbors' features through iterative message passing, building a comprehensive representation of the attack landscape. The GNN encodes each node into a numerical representation, positioning similar nodes close together in the embedding space. These learned embeddings support tasks like node classification, enabling the GNN to categorize unseen traffic instances into attack types.

### F. The attention mechanism

Inspired by recent advancements in computer vision tasks such as object detection [28], [29], the attention mechanism—particularly the multi-head attention mechanism [30]—has increasingly been integrated into NetFlow-based attack detection models [31], [32]. The attention mechanism is designed to selectively emphasize the most critical regions of an image while disregarding less relevant areas. It operates using a query-key-value framework, where the query (*Q*) represents the aspect requiring focus, the key (*K*) corresponds to input features potentially relevant to *Q*, and the value (*V*) contains the information associated with *K*. The mechanism computes a weighted sum of the values, with the weights determined based on the relevance of the keys to the query

### G. Model Training and evaluation

For model training and performance evaluation, the experimental dataset was randomly split into 80%-20% train-test sets. In addition, 10% of train-batch was randomly set aside for validation during each training epoch. For each of the dimension reduction methods the high dimensional train set was transformed into an 8-dimensional representation. For neuron activation, the ReLU activation function was used while Adam was employed for model optimization with a learning rate of 0.001. For each dimension reduction and graph data construction configuration the model was trained for 20 epoch using a batch size of 128.

## IV. RESULTS AND DISCUSSION

### A. Impact of n_neighbors-metric combination on detection performance

The attack classification accuracy of the model was evaluated for the three dimension reduction methods. In addition, for each method the accuracy was recorded for four *n_neighbors-metric* combinations. Fig. 5 provides a visual comparison of model performance for the three dimension reduction methods across the four *n_neighbors-metric* combinations.

The results clearly indicate that training the model on graph data constructed using the *Euclidean* metric generally outperforms the *cosine* metric. Furthermore, using 3 neighbors for graph data construction yields better performance compared to using 5 *neighbors*. Additionally, the VAE-encoder-based dimension reduction approach achieved the highest accuracy,

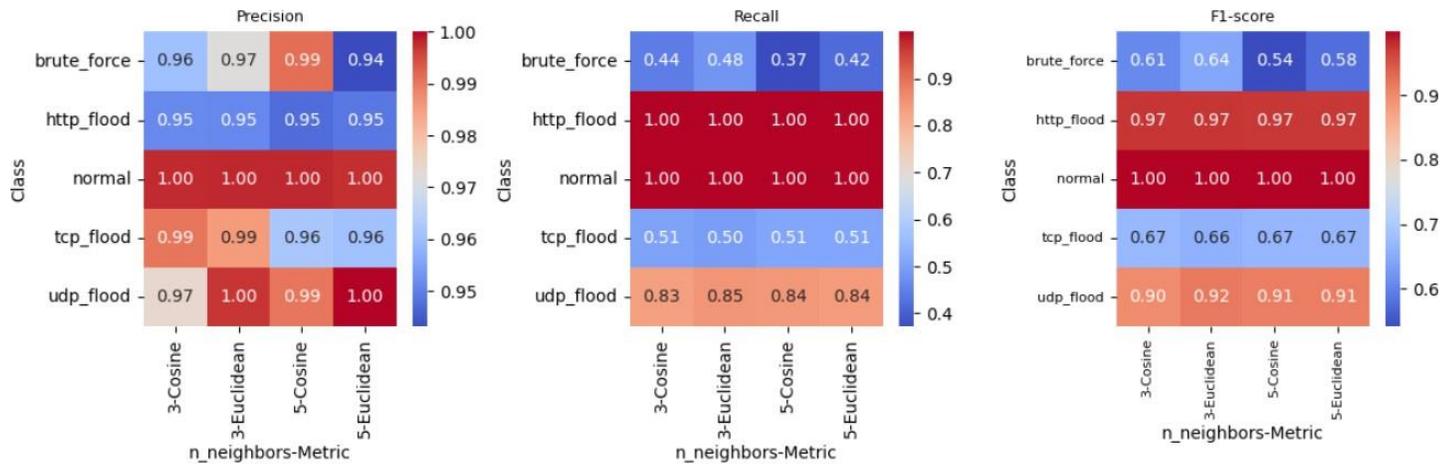

Fig. 2: Per-class model performance, in terms of precision, recall, and F1-score, for AE-based dimensional reduction and various *n_neighbors-metric* combinations for KNN-based graph data construction

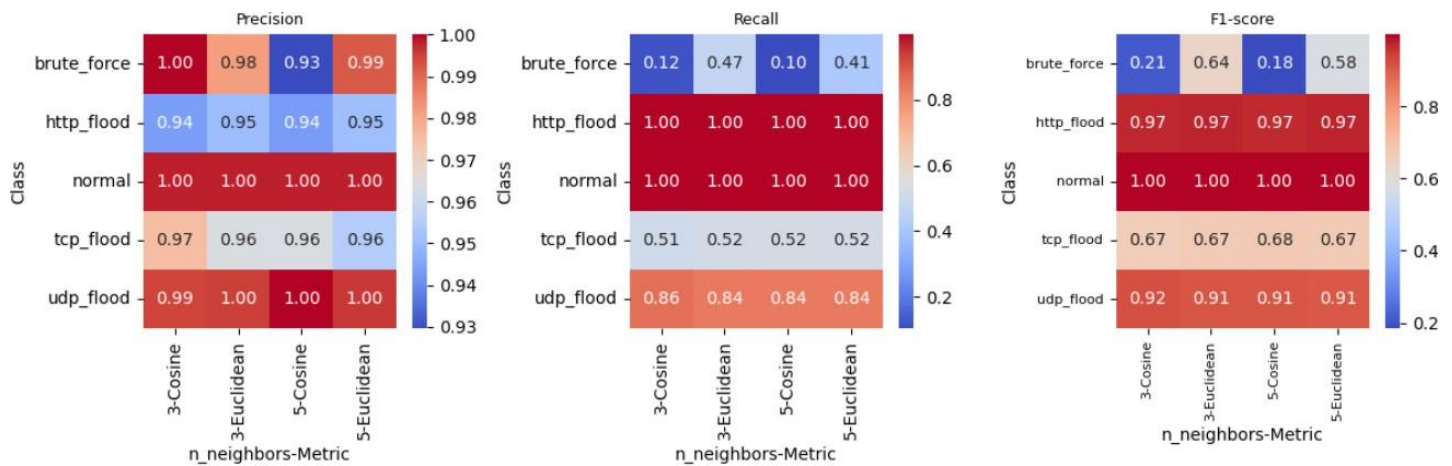

Fig. 3: Per-class model performance, in terms of precision, recall, and F1-score, for VAE-based dimensional reduction and various *n_neighbors-metric* combinations for KNN-based graph data construction

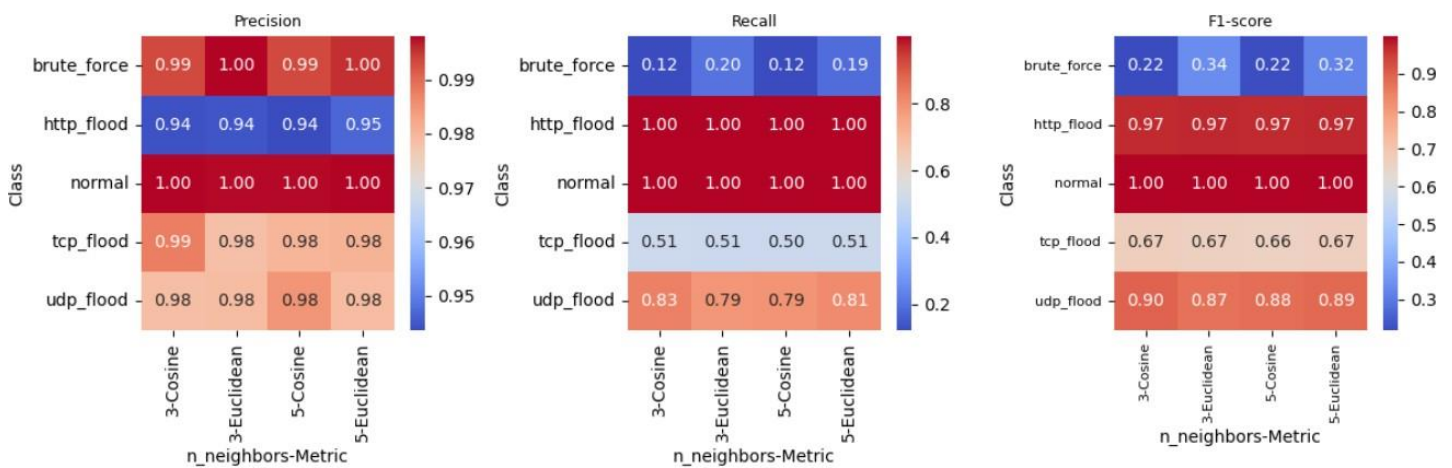

Fig. 4: Per-class model performance, in terms of precision, recall, and F1-score, for PCA-based dimensional reduction and various *n_neighbors-metric* combinations for KNN-based graph data construction

reaching 98.15% for the 3-Euclidean *n_neighbors-metric* combination and 98.12% for the 5-Euclidean *n_neighbors-metric* combination.

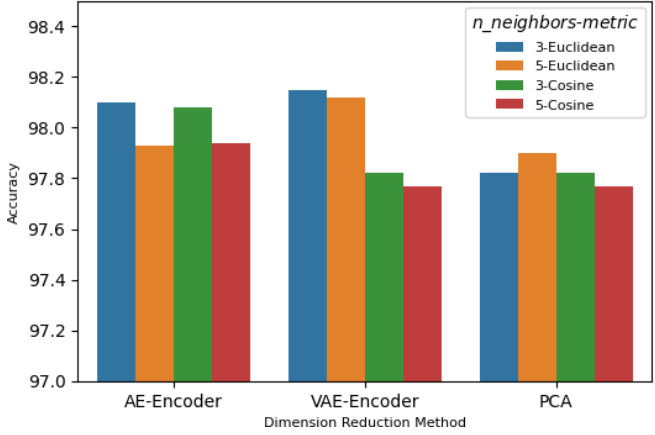

Fig. 5: Detection accuracy for the three dimension reduction methods across the four *n_neighbors-metric* combinations

### B. Per-class detection performance for the three dimension reduction across the four n_neighbors-metric combinations

To gain insight into the impact of each dimension reduction method on the recognition of individual traffic classes, a classification report in terms of precision, recall and F1-score was generated for each reduction method and results compared across the four *n_neighbors-metric* combinations. The results are present in Fig. 2 for the AE-based method while Fig. 3 and Fig. 4 present results for VAE-based and PCA-based dimension reduction methods, respectively.

It is clear from the results that, despite demonstrating high detection accuracy and competitive precision, the model records a low recall and F1-score for the "tcp_flood" traffic across all dimension reduction methods and *n_neighbors-metric* combinations. Similarly, for all configurations the model struggles to accurately recognize "brute_force" instances. This performance degradation can be partially attributed to class imbalance and the presence of errors commonly arising from mislabeled traffic instances [33]. A detailed investigation into the underlying causes is deferred to future work. In addition, PCA records the lowest performance across all *n_neighbors-metric* combinations. This can be attributed to the fact that, in addition to the limitations highlighted in Section III-C, the 8 principal components only account for 91.61% variance of the training dataset distributed as (40.97%, 19.79%, 9.94%, 8.58%, 4.68%, 3.00%, 2.78%, 1.87%).

### C. Computational cost

The computational cost in terms of time is a summation of the dimension reduction time, graph data construction time and model training time.

**AE and VAE Time Complexity**: Both AE and VAE use dense layers in their encoder and decoder models. A dense layer with input dimension $d_{in}$ and output dimension $d_{out}$ has a time complexity of $O(d_{in} \cdot d_{out})$. VAE additionally involves latent space sampling with a constant time complexity of $O(1)$. Thus, for $a$ encoder and $b$ decoder dense layers, AE and VAE have time complexities of $O(a \cdot d_{in} \cdot d_{out} + b \cdot d_{in} \cdot d_{out})$ and $O(a \cdot d_{in} \cdot d_{out} + b \cdot d_{in} \cdot d_{out} + 1)$, respectively. These can further be expressed as $O(c(\cdot d_{in} \cdot d_{out}))$ and $O(c(\cdot d_{in} \cdot d_{out}) + 1)$, respectively, where $c = a + b$.

**PCA time complexity**: With $N$ as the sample size, $D$ as the number of features, and $C$ as the number of principal components, the time complexity using iterative methods like randomized SVD is approximately $O(NDC)$, as only the first $C$ components are computed.

**Graph data construction time complexity**: The time complexity of constructing a sparse graph with $N$ nodes (each of dimension $D$) and $E$ edges is $O(ND^2 + ED)$.

**GAT model time complexity**: The time complexity of the GAT model with $n$ GATConv layers, $N$ nodes, $E$ edges, input dimension $D$, attention heads $H$, and output feature size $K$ per head is $O(n(NDK + HEK))$.

Table I summarizes the computational complexity associated with using each of the dimension reduction methods.

TABLE I: Computational cost per dimension reduction method

| Reduction method | Time Complexity |
| --- | --- |
| AE-encoder | $O(c(\cdot d_{in} \cdot d_{out})) + O(ND^2 + ED) + O(n(NDK + HEK))$ |
| VAE-encoder | $O(c(\cdot d_{in} \cdot d_{out}) + 1) + O(ND^2 + ED) + O(n(NDK + HEK))$ |
| PCA | $O(NDC) + O(ND^2 + ED) + O(n(NDK + HEK))$ |

From the results, the PCA-based framework demonstrates poorer detection performance compared to the AE- and VAE-based frameworks. However, the AE- and VAE-based frameworks incur significantly higher computational costs. Therefore, one must consider a trade-off between detection performance and computational cost in selecting which framework to deploy.

## V. CONCLUSION

IoT-based botnet detection models often neglect inter-instance relationships, limiting classification performance. Graph Neural Networks (GNNs) address this by embedding similar instances closer based on node features and relationships. However, transforming flow-based IoT traffic into graph structures introduces challenges such as computational overhead in terms of memory and processor requirements due to high dimensionality of instance nodes and the additional edge information. To mitigate these issues, this study proposed a framework that applies dimension reduction before graph data construction. Three dimension reduction techniques—AE-encoder, VAE-encoder, and PCA—were evaluated to assess their impact on a Graph Attention Neural Network (GAT) model for botnet attack detection, improving efficiency and classification accuracy.